\documentclass[pmlr]{jmlr}

\usepackage{longtable}
\usepackage{paralist,color}
\usepackage{caption}
\usepackage{subcaption}
\usepackage{float}
\usepackage{hyperref}
\usepackage{multirow}
\usepackage{array,ragged2e}

\usepackage{booktabs}
\usepackage[load-configurations=version-1]{siunitx} 

\theorembodyfont{\upshape}
\theoremheaderfont{\scshape}
\theorempostheader{:}
\theoremsep{\newline}

\jmlryear{2019}
\jmlrworkshop{Machine Learning for Healthcare}

\title[What Clinicians Want: Contextualizing Explainable ML for Clinical End Use]{What Clinicians Want: Contextualizing Explainable Machine Learning for Clinical End Use}

\author{\Name{Sana Tonekaboni}\textsuperscript{*}\textsuperscript{1,2,4}  \Email{stonekaboni@cs.toronto.edu} \\
\AND
\Name{Shalmali Joshi}\textsuperscript{*}\textsuperscript{2}  \Email{shalmali@vectorinstitute.ai} \\
\AND
\Name{Melissa D. McCradden}\textsuperscript{2,3,4} \Email{melissa.mccradden@sickkids.ca}\\ 
\AND
\Name{Anna Goldenberg}\textsuperscript{1,2,4}  \Email{anna.goldenberg@vectorinstitute.ai} \\
\AND
\addr \textsuperscript{1} Department of Computer Science\\University of Toronto, Toronto, Canada\\
\textsuperscript{2} Vector institute for artificial intelligence\\ Toronto, Ontario, Canada\\
\textsuperscript{3} Department of Bioethics\\The Hospital for Sick Children, Toronto, Canada\\
\textsuperscript{4} Department of Genetics \& Genome Biology\\ The Hospital for Sick Children, Toronto, Canada\\} 

\begin{document}

\maketitle

\begin{abstract}
Translating machine learning (ML) models effectively to clinical practice requires establishing clinicians' trust. Explainability, or the ability of an ML model to justify its outcomes and assist clinicians in rationalizing the model prediction, has been generally understood to be critical to establishing trust. However, the field suffers from the lack of concrete definitions for usable explanations in different settings. To identify specific aspects of explainability that may catalyze building trust in ML models, we surveyed clinicians from two distinct acute care specialties (Intenstive Care Unit and Emergency Department). We use their feedback to characterize when explainability helps to improve clinicians' trust in ML models. We further identify the classes of explanations that clinicians identified as most relevant and crucial for effective translation to clinical practice. Finally, we discern concrete metrics for rigorous evaluation of clinical explainability methods. By integrating perceptions of explainability between clinicians and ML researchers we hope to facilitate the endorsement and broader adoption and sustained use of ML systems in healthcare.
\end{abstract}

\section{Introduction}\label{sec:intro}

For clinical Machine Learning (ML), lack of model robustness~\citep{papernot2016transferability}, complexity of clinical modeling tasks~\citep{ghassemi2018opportunities}, and high stakes ~\citep{vayena2018machine} are some of the technical barriers to practical adoption. Additionally, even a highly accurate ML system is not necessarily sufficient in and of itself to be routinely utilized and endorsed by clinical staff~\citep{bedoya2019minimal,guidi2015clinician}. The uptake and sustained use of ML in healthcare has been more challenging than anticipated, as observed through piloting models for cardiac arrest~\citep{smith2013ability} and sepsis~\citep{masino2019machine,elish2018stakes}. 
This brings about the question of how truly reliable ML systems can earn clinicians' trust for sustained use and positive clinical impact. 

Currently, many rule-based assistive tools, such as different early warning scores~\citep{prytherch2010views, vincent1996sofa,duncan2006pediatric}
, have been deployed in clinical settings. By their very nature, rule-based algorithms are not considered opaque and in need to be endowed with auxiliary explanations of model outcomes. However, recently developed ML based tools that aim to improve on rule based systems are not considered to have the same level of accepted transparency. Therefore, understanding ML model behaviour beyond conventional performance metrics has become a necessary component of ML research, especially in healthcare. This has led to the development of ``interpretable" or ``explainable" machine learning models~\citep{gunning2017explainable} as a measure to overcome barriers of trust and adoption. Unfortunately, as it stands right now this is a rather ill-defined problem~\citep{lipton2016mythos}. Existing work suffers from significant criticism due to the lack of objective definitions of what would validate a model in terms of explainability in the context of clinical practice, while only limited works focus on evaluating the quality and usability of the proposed explanation methods for the target end user~\citep{doshi2017towards}. The ML community has largely resorted to developing novel techniques for model explanations~\citep{guidotti2018survey} that are often insightful only to the algorithmic experts~\citep{miller2018explanation}. Recent commentators have suggested that perhaps the 'explainability' literature as it stands requires a pragmatic lens to better suit the needs of the end users who are ultimately the ones who will decide if the model survives in the real world or not~\citep{paez2019pragmatic}.   

For the purpose of this manuscript, we define ``Explainability in ML for Healthcare" as a set of measurable, quantifiable, and transferable attributes associated with an ML system targeted for clinicians to calibrate model trust. 
Our goal is to synthesize a pragmatic notion of explainability for clinicians and effectively inform on--going research efforts in clinical ML at our institution. The integration of clinical ML in hospital settings is prompting the application of qualitative methodology in a novel context~\citep{elish2018stakes}. While previous work has assessed clinician use and acceptance of ML tools retrospectively and through quantitative surveys~\citep{guidi2015clinician}, we are among the first to investigate prospectively users' expectations of explainability. This method will enable us to anticipate potential challenges to translation as well as provide a baseline for future work to identify how explainability may evolve over time, in conjunction with the sustained use of ML tools. 

Additionally, we aim to better understand which ``explainability" methods can satisfy clinicians' needs. To this end, we conducted an exploratory pilot study with 10 clinical stakeholders to: 1) understand the role of `explainablity' for clinical ML system; 2) identify the classes of explanation that best serve the end user, emphasizing the context and diversity of their needs; 3) characterize how these considerations can be addressed from an ML standpoint; 4) propose  metrics for evaluating clinical explainability methods and identify gaps and research challenges in this area. 

\paragraph{Technical Significance}
We identify when and how explainability can facilitate reliable dissemination of ML model predictions in clinical settings. We curate a set of concrete classes of explanations based on the identified clinicians' needs.
We evaluate these in the context of existing explainability literature to highlight gaps and research challenges for clinical ML. 
By highlighting clinicians' needs for explanation as specific technical properties, we hope to channel explainable ML research intended for clinical end use to tackle some of the challenges.

\paragraph{Clinical Relevance}
Near future will likely  see increased adoption of ML solutions in key areas of clinical practice. Building trust between clinicians and ML models, for higher uptake and satisfaction requires concerted and purposeful efforts from the ML community. 
This work attempts to pre--emptively identify clinicians' needs,  via research surveys, to recognize concrete technical challenges that may be addressed using explainable ML with the objective of building reliable ML systems, fostering trust for sustained use aimed at improving clinicians' workflow and practice.



\section{Study Design}

We employed an upstream stakeholder engagement method~\citep{corner2012perceptions} to address the aforementioned questions. Upstream engagement here means that the survey is conducted prior to model implementation. Qualitative interviewing was determined to be an ideal method for the exploratory nature of this work, as surveys are insufficient for capturing the conceptual complexity of the phenomenon we want to understand. Furthermore, interviews provide an opportunity to clarify participants' responses and their reasoning. Interviews reached saturation~\citep{o2013unsatisfactory}, defined in the literature as the point wherein no new information pertaining to the main theme of explainability arose during the interviews which is generally expected after approximately 10 interviews~\citep{hennink2017code}. We developed an interview guide~\citep{jacob2012writing} to address the project goals. Conceptual development of `explainability' proceeded throughout the interviews~\citep{leech2009typology} to guide subsequent interview content~\cite{jacob2012writing}. 

We interviewed 10 clinicians in two acute care settings -- Intensive Care Unit (ICU) and Emergency Department (ED) with varying years of experience to develop notions of explainability and identify their needs towards building reliable ML systems for their respective clinical practice. These settings were chosen because both departments are areas where we foresee implementation of ML tools to support clinical care. Additionally, both departments have current experience working with either early warning/alert systems or opaque, non-ML decision support tools. The following describes the upstream stakeholder engagement method and study cohort in detail.


\subsection{Cohort}
We approached a convenience sample of stakeholder clinicians who were familiar with ML based clinical tools and are aware of ongoing developments in this field. Key stakeholders were defined as those who would be end--users of the technology and whose acceptance would ultimately determine successful clinical translation of predictive ML tools. Given the diversity of target clinical tasks, for use in the ICU and ED settings, 6 ICU clinicians and 4 ED clinicians were selected for this exploratory survey. These individuals (5 male, 5 female) were both senior (n = 5, completed residency training) and junior (n = 5, currently in residency training) clinicians to ensure the spectrum of relevant experience was represented among the stakeholders. Discussions were held at The Hospital for Sick Children, Toronto, Canada\footnote{\url{http://www.sickkids.ca/}} over a three week period at clinicians' convenience, and were recorded to facilitate recall accuracy.

\subsection{Procedures}
Clinicians' views were solicited to assess the concept of ML explainability with respect to their medical setting. Each interview lasted approximately 45-60 minutes. Our interviews began by exploring clinician's notions of `explainability' to determine what each clinician understood by the term and what he/she expected from ML models in their specific clinical setting. We then introduced the clinician to a hypothetical interactive scenario based on their specialties representing an ML tool incorporated into their division. For the ICU, we introduced a cardiac arrest (CA) prediction tool that reports risk of upcoming cardiac arrest~\citep{tonekaboni2018prediction}. For ED, we introduce a tool that predicts an acuity score based on triage reports~\citep{kwon2018validation}. We queried the range of possible responses that may arise when considering different aspects of model behavior. 


\subsection*{Hypothetical Scenarios}

ICU: Suppose a machine learning-based model has been integrated into the EHR system of the hospital. As vital measurements are being constantly recorded in the ICU, this model predicts the probability of a patient having a cardiac arrest within the next few minutes, based on temporal vital measurements. The risk score is updated every few minutes based on the patient’s current condition. Let us say a patient in your care is being monitored by the model and you receive an alarm that notifies you that they are at imminent risk of a cardiac arrest, but you had not suspected CA for a given patient at the moment.
\vspace{3mm}
\\
ED: A machine learning-based model has been integrated into the Emergency Department in your hospital. This system ranks incoming patients based on acuity using the information from the first triage. A patient in your care appears generally well, but the acuity score puts them in a high risk category.
\vspace{3mm}
\\
To enhance our understanding of the concept of explainability and trust as perceived by clinicians, further questions were asked to elucidate the following constructs: how would the clinician go about evaluating the validity of an alert in terms of translating it into clinically actionable information on the part of the patient?; how does this evaluation compare to other common non-ML tools, such as alerts, risk scores, or other notifications concerning a patient's status?; what do you currently do when one of these alerts/scores/notifications tells you something you weren't expecting about a patient?; what information would you want about this system? Additional questions are detailed in the Appendix.

\paragraph{Ethics Statement} Our initial exploratory investigation aligned with the Tri-Council Policy Statement's (TCPS2) (governing research-related activities in Canada) articles 6.11 and 10.1 stipulating activities exempt from Research Ethics Board approval. Activities exempted include: ``feasibility of the research, establish research partnerships, or the design of a research proposal". 
\section{Results}

We organize the results and analyses of the study as follows. 
We identify concrete and measurable set of explanation classes curated from the qualitative assessments of the surveys (Section~\ref{sec:results_translate_ml}). We identify a few metrics that help assess the utility of any class of explanation towards its clinical utility in Section~\ref{sec:res_metrics}. Finally, we summarize how existing explainable ML literature fares in the context of these clinical asks. A full synopsis of the qualitative data including illustrative quotes is provided in the Appendix.


\subsection{What makes a model explainable for clinicians?}\label{sec:results_translate_ml} 
Throughout the exploratory interviews it was clear that clinicians viewed explainability as a means of \emph{justifying} their clinical decision-making (for instance, to patients and colleagues) in the context of the model's prediction. To provide these explanations all clinicians expressed the need to understand the clinically relevant model features that align with current evidence-based medical practice. The implemented system/model needs to provide clinicians with information about the context within which the model operates and promote awareness of situations where the model may fall short (e.g., model did not use specific history or did not have information around certain aspect of a patient). \emph{Models that fall short in accuracy were deemed acceptable so long as there is clarity around why the model under-performs.} While it is ideal to learn models based on as much contextual information, it is not always practical and clinicians are aware of such a possibility. Furthermore, the complexity of clinical medicine is such that no model is likely to achieve perfect prediction; clinicians, in fact, expect this, and the acknowledgement of this challenge promotes trustworthiness. Clearly specifying features that go into the models for decision making, is a way to facilitate trust in the model as well as directing use in specific patient populations and determining parameters guiding appropriate use. The relevant quote -- `the variables that have derived the decision of the model'  was brought up by 3 ICU, 1 ED clinicians. This type of transparency is also identified in \cite{mitchell2018model} as being critical and needs to be disclosed to ML model users. While familiar metrics such as reliability, specificity, and sensitivity were important to the initial uptake of an AI tool, a critical factor for continued usage was whether the tool was \emph{repeatedly successful in prognosticating their patient's condition in their personal experience.} Real-world application was crucial to developing ``a sense of when it's working and when it's limited" which meant ``alignment with expectations and clinical presentation" [all clinicians].

Clinical thought process for acting on predictions of any assistive tool appears to consist of two primary steps following  presentation of the model's prediction: \begin{inparaenum} \item[i)] understanding, and \item[ii)] rationalizing the predictions. Thus classes of explanations for clinical ML models should be designed with the purpose of facilitating the understanding and rationalization process.
\end{inparaenum} Clinicians believe that carefully designed visualization and presentation can facilitate further understanding of the model. These features are essential to sustained model use largely due to the immediacy of the clinical picture being captured and in the context of multiple competing attentional demands that require user-friendly visualization. 
We discerned that there are situations where well-designed visualization is not enough and only additional explanation can facilitate and fill the gap in the identified clinical workflow.

\subsection{How Does this Translate to Reliable Clinical ML Design?}
We determine from our discussions that a well designed explanation should augment or supplement clinical ML systems to:
\begin{enumerate}[\hspace{0.5cm}(a)]\label{def:objectives}
\item Recalibrate clinician (stakeholder) trust of model predictions.
\item Provide a level of transparency that allows users to validate model outputs with domain knowledge.
\item Reliably disseminate model prediction using task specific representations (e.g. confidence scores).
\item Provide parsimonious and actionable steps clinicians can undertake. (e.g. potential interventions or data collection).
\end{enumerate}

We further curate the following classes of explanation from qualitative assessments that clinicians identified to most effectively complement model predictions based on our survey. We highlight wherever necessary, the applicability and importance of each explanation class for different settings (ICU vs ED). 


\paragraph{Feature Importance:}\label{subsec:feature_importance} Clinicians repeatedly identified that knowing the subset of features deriving the model outcome, is crucial. This allows them to compare model decision to their clinical judgment, especially in case of a discrepancy. In time-constrained settings such as the ED, important features are perceived as a crucial metric to draw the attention of clinicians to specific patient characteristics to determine how to proceed. While multiple clinicians mentioned the need to know relevant variables driving a prediction (see Appendix~\ref{app:quotes}), a junior clinician mentioned ``you have just a number, you can still use it but in your mind when you put all the variables that make you take a decision, the weight of that variable is going to be less than if you do understand exactly what that number means". It is also important to note that clinicians expect to see \emph{patient specific} variable importance as well as population level variable importance~\citep{james2013introduction,tibshirani1996regression}. While an extensive survey of feature importance is beyond the scope of this work (see Table~\ref{tab:gaps_in_literature} for a summary), we highlight that patient-level feature importance is a challenging and far less explored machine learning problem.

\paragraph{Instance Level Explanations:} Among commonly researched explainability methods, we investigated as part of our interviews whether data instances as explanations~\citep{kim2016examples,koh2017understanding} are useful in any clinical setting. Clinicians view this as finding similar patients~\citep{sun2012supervised,7837899,sharafoddini2017patient} and believe that this kind of explanation can be only helpful in specific applications. For example, in cases where an ML model is helping clinicians find a diagnosis for a patient, it is valuable to know the samples the model has previously seen~\citet{cai2019human}. However, in time constrained settings such as the ICU or ED, clinicians \emph{did not} find this explanation technique appealing. A non--trivial challenge with this class of explanation, that researchers need to be aware of, is the definition of similarity that is used~\citep{790428}. Clinicians identified that despite similar outcomes patients may differ significantly in the clinical trajectory to arrive of those outcomes, and vice versa. Various definitions for similarity may be proposed depending on the task the model is performing, and the way the user (clinician) wants to use the explanation. For example, some clinicians were interested in seeing similar samples to their patients because they were curious to know what actions were taken in those cases, and the associated outcomes of the interventions.

\paragraph{Uncertainty:} Clinicians overwhelmingly indicated that the model's overall accuracy was not sufficient and its clinical alignment in their judgment often determined their sustained use and trust in the model. That is to say that each time an alert presents a prediction, clinicians will anticipate a clinically significant change in the patient's status that should ideally align with the model's prediction. Previous reports have indicated that despite the determination of expert-agreed threshold for triggering alerts, there are still many alerts not accompanied by a clinically actionable change~\citep{umscheid2015development} which undermines use and endorsement by clinicians~\citep{guidi2015clinician}.
This point holds particular importance for clinical ML to ensure developments align with clinicians' expectations to promote trust and sustained model use. Presenting certainty score on model performance or predictions is perceived by clinicians as a sort of explanation that complements the output result. They also suggested that this score can be used as a threshold for reporting results only when they imply that model is very certain of its prediction. Many clinicians noted that ``alarm or click fatigue" (indicating the annoyance with repeating response prompts through the EHR system)~\citep{embi2012evaluating} is a significant concern that may be worsened by prediction tools. This issue is ubiqitous across healthcare contexts and requires careful consideration by model developers so as to not perpetuate clinician disillusionment and disengagement with these systems. This makes calibration of complex models~\citep{guo2017calibration} a significant technical challenge that needs to be addressed for clinical practice. 

An additional challenge is the fact that even models performing acceptably on average can have significant individual level errors~\citep{nushi2018towards} which is undesirable in clinical practice. There are two sources for uncertainty we highlight~\citep{gal2016uncertainty} that can affect model trust:
\begin{enumerate}
    \item Model Uncertainty: Uncertainty of model parameters that best explain observed data is known to be a significant source of uncertainty in model performance~\citep{nuzzo2014scientific,zhang2013domain}. Modeling for this uncertainty can be achieved by actively accounting for distributional differences during training~\citep{gal2016dropout, schulam2019auditing}. Few of these methods have been actively co-opted in clinical healthcare for explainability but nonetheless are crucial to uptake. 
    
    In addition, model mis-specification refers to performance deterioration due to a mis-specification of the model class used during training. It can lead to an overall underperforming system and can further reduce the effectiveness of some methods used for dealing with model uncertainty~\citep{wen2014robust}. This problem is especially under-studied for time-series models that heavily employ deep learning methods~\citep{zhu2017deep} and to the best of our knowledge remains mostly unexplored in clinical healthcare employing deep learning methods.
    
    \item Data Uncertainty: This type of uncertainty can come from noisy, missing data or an existing inherent uncertainty in the data. Both of these challenges are commonly faced in clinical ML. Inherently uncertain observation can be for instance a complicated patient with difficult diagnosis, thus is considered a relatively more difficult problem. Characterizing consistency under missingness has been only briefly studied in supervised classification system~\citep{josse2019consistency} and needs to be rigorously adopted and evaluated for clinical applications.
\end{enumerate}

\paragraph{Temporal Explanations:}  `Patient trajectories that are influential in driving model predictions' was reported to be an important aspect of model explanation. This area remains relatively unexplored in clinical ML explainability literature with a few exceptions~\citep{xu2018raim,choi2016retain, yang2018explaining} focused primarily on attention based and sensitivity analysis mechanisms in deep learning~\citep{vaswani2017attention}. However, explanations based on attention mechanisms can produce inconsistent explanations~\citep{jain2019attention}. Explainability for temporal data is a relatively challenging task as it requires analyzing higher order temporal dependencies. ML models that perform such predictions, should be able to explain their prediction based on changes in individual patient state. In units such as the ICU, clinicians are interested to see the change of state that has resulted in a certain prediction. Note that this is also related to explanations using Feature Importance (see Sub--section~\ref{subsec:feature_importance}) as a necessary component of providing explanations in the temporal domain. To the best of our knowledge, not much has been investigated in this domain, as formal definitions of explanations.\\

\paragraph{Transparent Design} Clinicians pointed to the need for models that reflect a similar analytic process to the established methodology of evidence-based medical decision making~\citep{haynes1997evidence}. They anticipate model decisions with similar emphasis on recognizing the clinical features driving the prediction as being more interpretable since it reflects how they currently assess a patient's risk status. Translating a model to a transparent design (e.g. a decision tree) is useful to facilitate rationalization of model behavior, as expressed by one ICU clinician: ``would want to know the equation to know what the weights are - but if it is variable and if knowing that is too much detail then it’s just not that helpful."[1 Senior ICU clinician] and ``if there is a discrepancy between the current clinical protocol and the model, and we don’t know why this is occurring, we are going to be nervous" [1 Senior ED clinician]. To the best of our knowledge, this is the most well studied area of model explanations in general and clinical ML literature. For example, a variety of rule based methods have been proposed precisely for the purpose of transparent clinical design~\citep{Lakkaraju:2016:IDS:2939672.2939874,wang2015falling}. Model distillation~\citep{hinton2015distilling} refers to learning a ``simpler" model class that perform at par with an complex ML model and can be considered as an attempt toward transparent design~\citep{che2015distilling}. Closely related are regularization based methods~\citep{wu2018beyond} that attempt to regularize models for enhanced interpretability. These methods tend to assume there are trade--offs between model performance and their perceived ``explainability". It is unclear when such a trade--off is acceptable, especially if it amplifies individual level errors. We further argue that such models should be rigorously evaluated for performance under other factors like mis-specification of the distilled model or regularized model, induced biases and general uncertainty to facilitate reliable adoption to clinical practice. \

\subsection{Metrics for Evaluating Explanations}\label{sec:res_metrics}

We further identify that explanations intended to build trust in clinical settings could benefit from being rigorously evaluated against the following metrics:

\paragraph{Domain Appropriate Representation} The quality of the explanation should be evaluated in terms of whether the representation is coherent with respect to the application task. For instance, for patients in the ICU, clinicians already have a lot of context as to why the patient is currently admitted to this division. Explanations that are redundant to the ICU task are not desirable unless critical to potential clinical workflows. The interviews revealed a diversity of settings in which clinicians operate, like diagnoses versus cardiac care management (see Instance based explanations in Section~\ref{sec:results_translate_ml}) which determines the representation best suited for appropriate settings. The representation itself should not further obfuscate model behavior for the clinician. This for instance requires careful filtering of the information most useful at any instant (e.g. highlighting ``age" even if it is an important feature is not useful if all potential clinical followups are uniform across age or it may not be helpful to receive an alert for a patient who is already en route to receiving a known intervention, like a surgery). The type of explanation will largely be evaluated by rigorous user studies involving stakeholders~\citep{doshi2017towards}. 

\paragraph{Potential Actionability} Given a model that satisfies minimal trust based criterion and has been sufficiently evaluated for performance, any complementary explanation should inform follow-up clinical workflow, including rationalization of model prediction (see Section~\ref{sec:results_translate_ml}). This followup can be anything ranging from checking on the patient to ordering additional lab measurements or determining an intervention. Overall, in an applied field like clinical ML, explanations that are informative (in terms of uncerstanding the mode), but have no impact on the workflow are of less importance. Similarly, the explanation should be parsimonious and timely. The nature of explanation should also account for inter-dependant factors that inform its usability. This can be determined by the purpose it will serve in any specific setting. For instance, ED clinicians believed that any risk assesment tool will help them allocate resources appropriately, therefore the explanations provided for the ML model should facilitate the decision making. For instance, in an ICU setting, when time to action is extremely limited, counterfactual explanations or patient similarity are not informative, usable, and result in cognitive overload~\citep{henriksen2008understanding}. 

\paragraph{Consistency} Consistency refers to two factors: \begin{inparaenum}\item[(i)] The set of explanations should be injective (or changes in model predictions should yield discernable changes in the explanation) \item[(ii)] These changes should be invariant to underlying design variations and should only reflect relevant clinical variability.  \end{inparaenum} Explanations that are inconsistent across any of these factors effectively violate their reliable actionability, and also negatively impact the trust of clinicians. Consistency is also closely related to robustness of explanations. For instance, such lack of consistency has already been identified using statistical methods for commonly used explanation techniques in deep learning literature~\citep{adebayo2018sanity,jain2019attention} outside of clinical ML.

\subsection{What is Missing in Explainable ML?}\label{sec:results_gaps}

While a multitute of novel explanability methods have been proposed in the  literature~\citep{guidotti2018survey}, they may not be directly applicable or have to be significantly tailored to clinical settings in the context of asks outlined in Section~\ref{sec:results_translate_ml}. In the following, we focus on identifying a collection of existing state of the art methods than can be used to generate explanations that are usable for clinical stakeholders. We summarize our results in Table~\ref{tab:gaps_in_literature} and identify possible shortcomings of existing ML approaches for clinical usability and relevance. Note that while a complete review of existing explainability methods is outside the scope of this work, we highlight the most relevant and popular techniques in the context of the classes outlined above.

Note that even if an existing explainability method is evaluated in a clinical setting, a rigorous evaluation of these methods for robustness and other inter--dependant factors still necessitates sustained research. Additionally, as noted in Section~\ref{sec:related_work} and Table~\ref{tab:gaps_in_literature}, some methods may work well for specific kinds of data, like images and may have to be non--trivially extended to be applicable to other data, and within the additional complexity of clinical settings.


\begin{table}[t]
\centering 
\caption{Summary of Explainable ML methods Contextualized for Clinical Applicability} 
\begin{tabular}{>{\RaggedRight\arraybackslash}p{3.5cm}|>{\RaggedRight\arraybackslash}p{5cm}|>{\RaggedRight\arraybackslash}p{5cm}}
\hline
Explanation Class &  Representative Existing Methods &  Possible shortcomings for clinical settings \\ \hline
\multirow{3}{3.5cm}{Feature Importance} & Sensitivity Analysis~\citep{saltelli2008global}, LRP~\citep{bach2015pixel}~\citep{yang2018explaining} &  Complex correlation between features of clinical models can be a challenge\\ 
& LIME, Anchors, Shapley Values~\citep{Ribeiro:2016:WIT:2939672.2939778,ribeiro2018anchors,lundberg2018explainable} & Further evaluation for consistency required\\
\hline
\multirow{3}{3.5cm}{Instance Level Explanation} & Influence functions~\citep{koh2017understanding} & Not evaluated on complex clinical models \\
& Prototypes and Criticisms~\citep{kim2016examples} & Limited applicability \\
\hline
\multirow{2}{3.5cm}{Uncertainty} & Distributional shift~\citep{subbaswamy2018counterfactual} & Not evaluated on complex clinical models\\  
& Parameter uncertainty~\citep{gal2016dropout,schulam2019auditing}&\\ 
\hline
Temporal Explanations & RETAIN, RAIM~\citep{xu2018raim,choi2016retain} & Potential lack of consistency due to the attention mechanism~\citep{jain2019attention}\\ 
\hline
Transparent  Design & Rule Based Methods~\citep{Lakkaraju:2016:IDS:2939672.2939874,wang2015falling} & Less powerful in modeling more complex applications; Generally assume a trade--off of accuracy and explainability\\
\hline
\end{tabular}
\label{tab:gaps_in_literature}
\end{table}
\section{Related Work}\label{sec:related_work}

To the best of our knowledge, no prior works have conducted target stakeholder studies for the specific purpose of identifying explainability challenges in clinical ML. In terms of translating ML methods to clinical practice,~\citet{escobar2016piloting,elish2018stakes} have extensively recorded their observations in piloting an early sepsis warning system to a single institute over a period of multiple days. A few user studies have evaluated prototypical Electronic Health Recording systems~\citep{mazur2016toward} or associated visualization prototypes~\citep{ghassemi2018clinicalvis} for efficacy of facilitating diverse clinical workflows~\citep{nolan2017multisite}. Herein we highlight some state of the art literature in explainable ML which we have discussed in terms of their relevance to adoption of ML models in clinical practice in Section~\ref{sec:results_translate_ml}.

Explainability in general machine learning has focused on understanding model behavior from different perspectives. Although simpler model classes~\citep{lundberg2018explainable,wu2018beyond} are considered more interpetable by users, a general trade-off between model performance and explainability is assumed~\citep{caruana2015intelligible}. Some methods like LIME and Anchors~\citep{Ribeiro:2016:WIT:2939672.2939778, ribeiro2018anchors} use simpler model classes to derive feature level explanations~\citep{sundararajan2017axiomatic}. Abstractions such as data instances~\citep{koh2017understanding,kim2016examples}  are used as a means of explaining model behaviors. Visualization is also explored as a method of providing an insight in model behaviour~\citep{yosinski2015understanding,ming2017understanding}. We characterize these methods as \emph{diagnostic} methods as they are not necessarily targeted to provide explanations for the target user but focus on improving model understanding. 
See ~\citet{guidotti2018survey} for an extensive survey of existing methods for explainable ML in general.\ 

In clinical ML, visualization systems have been evaluated for efficacy of interaction between EHR systems and healthcare practitioners~\citep{ghassemi2018clinicalvis}. Neural attention mechanisms~\citep{xu2015show} are used for developing interpretable models for heart failure and cataract on-set prediction~\citep{kwon2019retainvis, choi2016retain}.~\citet{wu2018beyond} use tree-based regularization for sepsis and HIV treatment, while~\citet{wang2015falling,ustun2016supersparse} build rule based or sparse linear methods for predicting hospital readmissions and sleep apnea scoring systems.~\citet{ahmad2018interpretable} conduct the most extensive survey of explainable methods in clinical healthcare, and analyze them in the context of existing notions in general ML and how they relate to clinical healthcare, discussing ethical questions thereof. They conclude that general agreement in the field of clinical healthcare toward explainability is fairly low. Our work attempts to bridge some of this uncertainty by initiating a conversation with and accounting for the needs of stakeholders.


Several works have attempted to codify rigorous evaluation of explainability methods. For instance,~\citet{doshi2017towards} propose ways to evaluate interpretability methods depending on application type and thereby the nature of evaluation, while~\citet{narayanan2018humans} and~\citet{poursabzi2018manipulating} conduct user studies to understand the quality of many proposed explainability methods. Statistical methods have also been introduced for evaluating the robustness of a few candidate explanation methods like saliency and neural attention architectures~\citep{adebayo2018sanity,jain2019attention}.


\section{Discussion}\label{sec:discussion}

This work documents the ongoing challenge of translation of clinical ML with a particular focus on explainability through the eyes of end users. We demonstrate how clinicians' views sometimes differ from existing notions of explainability in ML, and propose strategies for enhancing buy-in and trust by focusing on these needs. 
In light of the objectives highlighted in Section~\ref{sec:intro}, we survey clinicians with diverse specialties and identify when explainability methods can assist in enhancing clinicians' trust in ML models. Our research survey involved creating hypothetical scenarios of deploying a machine learning based predictive tool to carry out specific tasks in the ICU and the ED respectively. 
We demonstrate that by accounting for target stakeholders, even though the explainability task in clinical settings is significantly diverse, but it can be codified into specific technical challenges. We further outline the need to evaluate clinical explainability methods rigorously under the proposed metrics in Section~\ref{sec:res_metrics}.
To the best of our knowledge, this is the first attempt at involving ICU and ED stakeholders to identify targeted clinical needs and evaluating it against general machine learning literature in this field.

Some of our non--technical observations, as highlighted in the qualitative synopsis in Section~\ref{sec:results_translate_ml}  and Appendix~\ref{app:quotes} are corroborated to some extent by the observations of~\citet{elish2018stakes} who followed the development and deployment of a machine learning based sepsis risk detection tool in a clinical setting. Our work however is far more general, as we concretely map conclusions from clinical surveys to prominent gaps in explainable ML literature as it pertains to effective clinical practice. Limitations of our methods are highlighted below.

\subsection{Limitations and Future work}

Our research survey for identifying explainability challenges was restricted to ICU and ED specialists. Most specialists we interviewed have reasonable knowledge of clinical ML systems (aware of or involved in academic research involving clinical ML systems). Given the research questions which were focused on prediction tasks, the identified challenges may only have limited applicability to wider class of ML models (for example, reinforcement learning,  survival models etc). Applicability of the identified classes of explanations to other clinical settings (like outpatient tasks) needs further evaluation. While our study was an exploration into the research design of a broader study of explainability, we were able to glean useful insights.

A strength of the study is that we attempted to elucidate concepts under the explainability framework that can apply to many novel ML-based implementations. A larger research question we are currently pursuing is whether satisfaction of these explainability requirements is directly related to perceived usefulness of the tool as well as to sustained use and endorsement. We hope to develop a conceptual framework that can be utilized across a spectrum of ML tools that can support consistency and quality of ML's explainability requirements in hospital environments.


\acks{We would like to thank all the clinicians who volunteered their valuable time for this exploratory study.}

\bibliography{mlhc_survey}

\newpage
\appendix
\section{Interview Protocol}\label{app:protocol}

\subsection*{Additional Questions}

These questions were asked to get clinicans' perspective regarding specific explainability features (common in existing explainable ML literature as curated by the authors). We additionally asked questions that allowed us to get a sense of  potential metrics clinicians trust and would like the model to be supplemented with.

\begin{enumerate}
\item What kind of information could help you feel confident in how you chose to manage a patient?
\item What kinds of things do you look to now, for example, to help you estimate how a patient will do?
\item If you know the confidence of the model’s prediction for a specific instance, would that make any difference in your reaction? 
\item If the model presented to you the part of the patient measurements and history that has influenced the model decision the most, would you find it helpful? Do you think it could help save time when making a decision on interventions?
\item Do you think it would be helpful to get examples of similar patient with the similar trajectory? If so, what would a similar case look like for you? How would you use this information? Do you want to see similarity of clinical outcome or trajectory?
\item Are the set of top measurements that are driving the prediction useful?
\item Given all the points that were brought up (enumerate all options) , if you were back in the ICU/ED scenario, which one of the above added information would you find more useful and actionable?
\end{enumerate}

\section{Qualitative Synopsis and Representative Quotes from Stakeholders}\label{app:quotes}

Here we include representative quotes from stakeholders as evidence that formed the basis of our results. We present these in terms of general perception of clinicians towards ML systems as well as their perception of explainability.

\paragraph{General Perception of the Need for Clinical ML Systems:} Clinical ML systems that could support the improvement of patient outcomes were viewed as an important adjunct to facilitate care in both ICU and ED settings. 
Specific needs identified included the ability to have a system as a constant surveillance (e.g., collecting and analyzing physiologic signals), which was likened to having ``an extra tool in [their] toolbox" or a ``failsafe" [3 ED, 2 ICU, mostly junior], i.e. a tool that will be constantly monitoring the situation and alerting clinicians to the onset of potentially critical events. Several clinicians noted that the ability to intervene earlier was important in order to save time or prevent or minimize bad outcomes (e.g., loss of functionality, death). Successful ML implementations in this regard were likened to a senior clinician who ``picks up every sign so well" [3 ICU, mostly junior], in that the ML systems' value lies in its ability to mimic the clinical acuity of experienced specialists. 

Perhaps the most important benefit of ML was the ability to facilitate a risk- or acuity-based allocation of attention in the ED. Clinicians in the ED setting particularly described the challenge of not being able to predict with certainty the trajectory of individual patients, and interventions were thus necessarily reactive and subsequent to a negative clinical condition (e.g., blood pressure dropping suddenly). ML should be like having an ``constant eye on the patient'' [all clinicians] that could support the direction of appropriately timed resources to the bedside of a patient prior to the point of clinical decline. Supporting resource allocation at the precise moment of need was particularly beneficial in the ED, where ``there are a thousand things going on at once and you have incomplete information" [all ED clinicians]. In supporting appropriate attention allocation, clinicians refer to the ability of the ML model to provide a quick, reliable piece of information that clinicians can trust to direct them to the right patient at the right time.

\paragraph{Clinical Perception of Explainability of ML Systems:}
Primarily, clinicians viewed explainability as a means of \emph{justifying} their clinical decision-making (for instance, to patients and colleagues) in the context of the model's prediction. To them, the term often implied awareness of ``the variables that have derived the decision of the model" [3 ICU, 1 ED]. One senior clinician noted, ``if it's on a computer then the assumption is that somebody had already figured it out," implying that a lot of features may not require lengthy explanations but instead would require clinical accuracy, as we described in Section~\ref{sec:results_translate_ml}, (Uncertainty).
Explanations seemed to be closely tied to visualization and representation of model predictions. A few mentioned it might be helpful to know ``parameters that are feeding the model... what data is processed and aggregated" [1 Junior ICU clinician]. Clinicians also noted ``if there is a discrepancy between the current clinical protocol and the model, and we don’t know why this is occurring, we are going to be nervous" [1 Senior ED clinician]. Universally, awareness of the factors driving the prediction were viewed as essential to determining whether to do additional tests, return to bedside, or intervene: ``you have just a number, you can still use it but in your mind when you put all the variables that make you take a decision, the weight of that variable is going to be less than if you do understand exactly what that number means" [1 Junior ICU clinician]. Lastly, for sustained improvement using clinical feedback, knowledge of the model's development was perceived as important insofar as it facilitated the refinement of the tool itself. For example, one clinician noted it is ``crucial for clinicians to understand how the model is making predictions to be able to provide feedback on specific conditions that can influence the false positive rate" [1 Junior ED clinician]. 



\end{document}